\documentclass{article}
\usepackage{ijcai20} % The file ijcai20.sty is NOT the same than previous years'

\usepackage{times}
\usepackage{soul}
\usepackage{url}
\usepackage[hidelinks]{hyperref}
\usepackage[utf8]{inputenc}
\usepackage[small]{caption}
\usepackage{graphicx}
\usepackage{amsmath}
\usepackage{amsthm}
\usepackage{booktabs}
\usepackage{algorithm}
\usepackage{flushend} %can coz problems like lines overlap

\usepackage{amssymb}
\usepackage{algpseudocode}
\algnewcommand{\Inputs}[1]{%
  \State \textbf{Input:}
  \Statex \hspace*{\algorithmicindent}\parbox[t]{.8\linewidth}{\raggedright #1}
}
\algnewcommand{\Initialize}[1]{%
  \State \textbf{Initialize:}
  \Statex \hspace*{\algorithmicindent}\parbox[t]{.8\linewidth}{\raggedright #1}
}
\algnewcommand{\Outputs}[1]{%
  \State \textbf{Output:}
  \Statex \hspace*{\algorithmicindent}\parbox[t]{.8\linewidth}{\raggedright #1}
}
\usepackage{color}
\definecolor{Blue}{rgb}{0,0,1}

\definecolor{Orange}{rgb}{1,0.5,0}

\definecolor{Green}{rgb}{0,1,0}

\title{Improving Label Ranking Ensembles\\ using Boosting Techniques}

\author{
Lihi Dery$^1$
\and
Erez Shmueli$^2$ 
\affiliations
$^1$Ariel University\\
$^2$Tel Aviv University
}

\begin{document}

\maketitle

\begin{abstract}
Label ranking is a prediction task which deals with learning a mapping between an instance and a ranking (i.e., order) of labels from a finite set, representing their relevance to the instance.
Boosting is a well-known and reliable ensemble technique that was shown to often outperform other learning algorithms.
While boosting algorithms were developed for a multitude of machine learning tasks, label ranking tasks were overlooked.
In this paper, we propose a boosting algorithm which was specifically designed for label ranking tasks. 
Extensive evaluation of the proposed algorithm on 24 semi-synthetic and real-world label ranking datasets shows that it significantly outperforms existing state-of-the-art label ranking algorithms.
\end{abstract}

\section{Introduction}
Label ranking is a prediction task which deals with learning a mapping between an instance and a ranking (i.e., order) of labels from a finite set, representing their relevance to the instance \cite{zhou2014taxonomy}.
%
% Label ranking is the problem of predicting and ranking multiple labels for a single instance \cite{vembu2010label,zhou2014taxonomy}.
% Each instance has several predictive attributes and a target value that contains a (possibly incomplete) ranking of labels.
%
Due to its wide applicability, label ranking has attracted a lot of focus from the artificial intelligence community in recent years \cite{hullermeier2008label,cheng2009decision,aiguzhinov2010similarity,cheng2013labelwise,zhou2014taxonomy,gurrieri2014alternative,destercke2015cautious,aledo2017tackling,sa2017label,ZHOU2018}.
%
%This is in part due to the large number of applications in which it is important or desirable to order the instance's labels. 

%Label ranking is abundant in various applications in which it is important or desirable to order the instance's labels. 
Applications of label ranking include for example, text classification, where a news article may belong to multiple topics, and the goal of the label ranking algorithm is
to rank the topics according to their relevance to the document.
In pattern recognition, objects can be ordered according to their relevance to the image ~\cite{yang2016exploit}.
In meta-learning, a label ranking model can provide a list of algorithms to a given problem, ranked according to their fit to the problem, based on the characteristics
of the problem at hand \cite{brazdil2003ranking}.
%Recommendation problems can also be designed as label ranking tasks, where each user is an instance. For example, recommending YouTube videos (the labels) \cite{weston2013label}, or predicting which ads (the labels) are most fit for display to a user \cite{djuric2014non}.

Label ranking tasks must not be confused with multi-label tasks \cite{tsoumakas2009mining} nor with learning to rank tasks \cite{cohen1998learning}. 
In label ranking tasks, the target attribute of each instance contains a ranking of labels, representing their relative relevance to the instance. 
%In label ranking problems, the goal is to predict the correct order (ranking) of labels for each instance, out of a finite and predefined set of labels.
In contrast, in multi-label tasks, the target attribute is a non-ranked subset of relevant labels, and in learning to rank tasks, the goal is to produce a ranked list of the instances themselves.

Boosting is a well-known and reliable ensemble technique \cite{Schapire2003} that was shown to often outperform other learning algorithms.
AdaBoost \cite{freund1997decision} is one of the most widely used classification boosting techniques.
Variations of AdaBoost were developed for 
multi-class tasks \cite{freund1997decision},
multi-label tasks \cite{schapire2000boostexter},
regression tasks \cite{drucker1997boosting,solomatine2004adaboost}, 
and learning to rank tasks \cite{cohen1998learning,xu2007adarank,wu2010adapting}. 
However, to the best of our knowledge, no boosting algorithm was suggested for label ranking tasks.

%\textbf{Contibutions:}
In this paper, we propose a novel boosting algorithm, AdaBoost.LR, which was specifically designed for label ranking tasks. 
An extensive evaluation of AdaBoost.LR over 24 semi-synthetic and real-world datasets shows that it significantly outperforms existing state-of-the-art label ranking algorithms.

The rest of this paper is organized as follows: 
we first discuss label ranking ensembles in section \ref{sec:related_work}.
Thereafter, we describe our proposed method in section \ref{sec:proposed_method}.
This is followed by an overview of our experimental setting in section \ref{sec:experimental_setting}, a description of our extensive evaluation in section \ref{sec:results}, and a summary and suggestions for future work in section \ref{sec:conclusions}.

\section{Related Work}
\label{sec:related_work}

Numerous label ranking algorithms were suggested in the literature. 
One approach is based on turning the problem into several binary classification problems and then combining them into output rankings.
(e.g. \cite{hullermeier2008label,cheng2013labelwise,destercke2015cautious}).
Another common approach is based on modifying existing probabilistic algorithms to directly support label ranking. Some main examples are: 
naive Bayes models \cite{aiguzhinov2010similarity}, 
k-nearest neighbor models \cite{brazdil2003ranking} 
and decision tree models \cite{cheng2009decision,de2015distance}.
%Label Ranking Trees (LRT) \cite{cheng2009decision} and Entropy Based Ranking Trees (ERT) \cite{de2015distance}.
For a surveys on label ranking algorithms see e.g. \cite{zhou2014taxonomy}.

In order to improve the performance of label ranking algorithms, four recent papers suggested the use of ensembles
\cite{aledo2017tackling,sa2017label,werbin2019beyond,ZHOU2018}.
The ensembles proposed in these papers differ in several aspects.
% (1) the base label ranking algorithm used,
% (2) the method used to sample the data to train each of the simple models (if all models are trained with the exact same data they will output the exact same results, and then there is no need for an ensemble), and
% (3) the aggregation method used to combine the results of the simple models.
Aledo et al. \shortcite{aledo2017tackling} used Label Ranking Trees (LRT) \cite{cheng2009decision} as the base label ranking algorithm , whereas Sa et al. \shortcite{sa2017label} used Ranking Trees (RT) and Entropy Ranking Trees (ERT) \cite{de2015distance}, Zhou and Qiu. \shortcite{ZHOU2018} developed their own method named Top Label As Class (TLAC) and Werbin et al. \shortcite{werbin2019beyond} used both LRT and Ranking by Pairwise Comparison (RPC)  \cite{hullermeier2008label}.
To select the training data for each simple classifier, Aledo et al. \shortcite{aledo2017tackling} and Werbin et al. \cite{werbin2019beyond} used a technique known as Bootstrap aggregation or Bagging \cite{breiman1996bagging}, whereas Sa et al. \shortcite{sa2017label} and Zhou and Qiu \shortcite{ZHOU2018} suggested modifications to the well-known Random Forest technique \cite{breiman2001random}.
As for the aggregation method used, three studies \cite{de2015distance,aledo2017tackling,ZHOU2018} used a single predefined voting rule (either Borda or Modal Ranking) and one study \cite{werbin2019beyond} used a Voting Rule Selector (VRS).  

%Boosting algorithms proceed in rounds; at each round, a subset of the training data is fed to the learner.
% In the first round, the training sample is chosen at random.
% In the proceeding rounds, samples on which the previously learned model erred are given a higher weight and thus have a higher probability to be selected.
% After a predetermined number of rounds, the models' outputs are combined, usually by means of weighted majority voting. 

To the best of our knowledge, boosting for label ranking has been so far overlooked. The boosting variations that are perhaps the most closely related to the label ranking task are AdaBoost.MR \cite{schapire2000boostexter}, which was designed for multi-label tasks, and AdaBoost.R2, which was designed for regression tasks \cite{drucker1997boosting}. 
%AdaBoost.MR performs pairwise comparisons between labels. Training instances and their corresponding label pairs that are hard to predict, receive incrementally higher weights in following classifiers while instances and label pairs that are easy to classify receive lower weights. Pairs of labels that are hard to predict are defined as pairs where one label belongs to the target class and the other does not. 
%As label ranking is not the goal of this algorithm, the algorithm is designed to minimize the ranking loss, i.e., the number of misordered pairs where a irrelevant label is classified as relevant and an relevant label is classified as irrelevant. Furthermore,
AdaBoost.MR assumes that each instance in the training set is associated with a subset of relevant labels, and given a new instance, the goal is to predict a subset of relevant labels.
AdaBoost.R2 assumes that each instance in the training set is associated with a numeric value, and given a new instance, the goal is to predict a numeric value.
In contrast, our proposed boosting-based algorithm assumes that each instance in the training set is associated with a ranking (not necessarily complete) of all labels, and given a new instance, the goal is to predict a complete ranking of all labels.
We proceed to describe the proposed algorithm in section \ref{sec:proposed_method}.

\section{The proposed method}
\label{sec:proposed_method}
We now describe our proposed boosting algorithm for label ranking tasks: $AdaBoost.LR$.
A pseudo-code of the algorithm is provided in Algorithm \ref{alg:AdaBoost.LR}.

\begin{algorithm*}[h]
	\caption{AdaBoost.LR Algorithm}
	\label{alg:AdaBoost.LR}
	\begin{algorithmic}[1]
		\Inputs{A set of $m$ training instances, $D$, with rankings as targets\\
		A base label ranking learning algorithm $weak\_learner$ \\
		The maximum number of iterations $T$\\
		The sample ratio $S$}
	    \Initialize{Iteration $t=1$\\
	    Weights: \strut$w_1(i) \gets \frac{1}{m}$, \quad $1 \leq i \leq m$ \\ 
	    Average loss $\bar{L_1} \gets 0$}
	    \While{$\bar{L_1}<0.5 \  and \ t<=T $}
	       \State $D_s \gets$ sample an $S$ fraction of instances from $D$ based on their weights $w_t$  
          \State Call $weak\_learner(D_s)$ and get a hypothesis $f_t(x) \to y$
          \State Calculate the loss $l_t(i)$ for each training instance $i$:  $l_t(i)= 1-kt(f_t(x_i), y_i)$
          \State Calculate the adjusted loss $L_t(i)$ for each training instance $i$: $L_t(i)=\frac{l_t(i)}{\smash{\displaystyle\max_{1 \leq i \leq m}} l_t(i) }$
          \State Calculate the average loss: $\bar{L_t}= \sum\limits_{i=1}^{m} L_t(i)w_t(i)  $
          \State Calculate the model's confidence: $\beta_t= \frac{\bar{L_t}}{1-\bar{L_t}} $
            \State Calculate the model's weight: $\alpha_t=  \log(\frac{1}{\beta_t}) $
          \State Update the weight $w_{t+1}(i)$ for each training instance $i$:  $w_{t+1}(i) = \frac{w_t(i) \cdot \beta_t^{1-L_t(i)}}{\sum\limits_{i=1}^{m} w_t(i) \cdot \beta_t^{1-L_t(i)}}$
          \State Set $t=t+1$
        \EndWhile
	    \Outputs{ 
	    Weak models $f_t$ and their weights $\alpha_t$\\
	    Final hypothesis - given a new instance, aggregation of the weak models' outputs is done using weighted Borda:  %\setlength\parindent{24pt}
	    $score(c_i)=\sum\limits_{t=1}^{T} \sum_{j=1,j \neq i}^n N_t(c_i,c_j)\cdot\alpha_t$, where all labels $c_i$ are sorted by their scores (decreasing order).
	    }
	\end{algorithmic}
\end{algorithm*}

The algorithm receives as input a training set of $m$ instances with rankings as targets, a base label ranking learning algorithm, and the maximum number of iterations $T$ (line 1). 
The algorithm begins by initializing the weights of all training instances to the same value (line 2).

\begin{equation}
\label{eq: first_weights}
w_1(i)=\frac{1}{m}
\end{equation}

\noindent
Then, the algorithm proceeds in iterations as follows (lines 3-13).
First, a set of training instances $D_s$ is sampled based on the current weights $w_t$ of the instances (line 4)..
Next, the base learning algorithm is called with the sampled set of training instances, and the resulting trained model (hypothesis) $f_t(x) \to y$ is returned (line 5).
Then, the the trained model is applied on each training instance, and the loss for each training example, denoted by $l_t(i)$, is calculated as the Kendall-tau-b coefficient \cite{kendall1938new}, $kt$, between the predicted ranking $f_t(x_i)$ and the target ranking $y_i$ of training instance $i$ (line 6):

\begin{equation}
\label{eq:loss_by_kt}
l_t(i)= 1-kt(f_t(x_i), y_i)
\end{equation}

\noindent
Next, the adjusted loss for each training instance $i$, $L_t(i)$, is calculated by dividing its loss $l_t(i)$ by the maximum loss of all training instances (line 7): 

\begin{equation}
\label{eq: inst_loss normalized}
L_t(i)=\frac{l_t(i)}{\smash{\displaystyle\max_{1 \leq i \leq m}} l_t(i)}
\end{equation}

\noindent
The performance of the weak model is then evaluated by computing the average (adjusted) loss over all instances (line 8):

\begin{equation}
\label{eq:avg_loss}
\bar{L_t}=\sum\limits_{i=1}^{m} L_t(i)w_t(i) 
\end{equation}

\noindent
In the next steps, the confidence of the model $\beta_t$ (line 9) and the weight of the model $\alpha_t$ (line 10) are calculated: 

\begin{equation}
\label{eq:beta}
\beta_t = \frac{\bar{L_t}}{1-\bar{L_t}} 
\end{equation}

\begin{equation}
\label{eq:alpha}
\alpha_t = log(1/ \beta)
\end{equation}

\noindent
Then, the weights of the training instances are updated towards the next iteration. 
Training instances for which the model predicted poorly will receive a higher weight, thereby "forcing" the next trained model to "pay special attention" to them. 
The calculation of the new weight, $w_{t+1}(i)$, for a training instance $i$, depends on the confidence of the model at time $t$, ($\beta_t$), and the adjusted similarity, $(1-L_t(i))$, between the predicted value and the actual target (line 11):

\begin{equation}
\label{eq:inst_weigth}
w_{t+1}(i) = \frac{w_t(i) \cdot \beta_t^{1-L_t(i)}}{Z_t}
\end{equation}

\noindent
Where $Z_t$ is a normalization factor: $Z_t = \sum\limits_{i=1}^{m} w_t(i) \cdot \beta_t^{1-L_t(i)}$.
The process is repeated until $T$ weak models are constructed or until $\bar{L_i} \geq 0.5$ (line 3).
Finally, the algorithm outputs the weak models $f_t$ and their weights $\alpha_t$ (line 14).

\bigskip
Prediction for a new (unseen) instance is done by retrieving the weak models' outputs and their weights, and then employing weighted Borda.
Formally, let there be $n$ labels $C=\{c_1, c_2,\ldots,c_n\}$.
Let $N_t(c_i,c_j)$ be $1$ if $c_i$ was ranked above $c_j$ in the output of model $t$, and $0$ otherwise.
The score of label $c_i$ is:

\begin{equation}
\label{eq:weighted_borda}
score(c_i)=\sum\limits_{t=1}^{T} \sum_{j=1,j \neq i}^n N_t(c_i,c_j)\cdot\alpha_t
\end{equation}

\noindent
The $n$ labels are then ranked according to their scores.

\flushcolsend

\newpage
Our proposed boosting algorithm for label ranking tasks, AdaBoost.LR, was inspired by the AdaBoost.R2 algorithm for regression tasks \cite{drucker1997boosting}.
Notable differences between AdaBoost.R2 and AdaBoost.LR are highlighted below:

\begin{itemize}

\item \textbf{Nature of target attribute and base algorithm} - In regression tasks the target attribute contains numeric values.
In contrast, in label ranking tasks, the target attribute contains rankings of labels from a predefined set.
Consequently, the base algorithm used by a label ranking ensemble must be able to handle a label ranking target.

\item \textbf{Loss calculation} - In regression tasks, the loss is computed as the absolute difference between the predicted 
and actual numeric values.
However, in our case, we need to calculate the error between the predicted and actual rankings.
To do so, we used the Kendall-tau coefficient function.
Kendall-tau counts the minimal number of swaps needed to transform one ranking into another ranking.
Specifically, we used Kendall-tau-b which can handle ties in the order of the two rankings, and a normalized version which measures the similarity between two rankings on a scale of $[-1,1]$ ($-1$ means that the two rankings are entirely opposite and $1$ means the two rankings are identical). 
Thus, equation \ref{eq:loss_by_kt} measures (for each training instance $i$ at time $t$) the difference between the predicted ranking and the real ranking.
A training instance $i$ with high $l_t(i)$ value means it was poorly predicted.

\item \textbf{Output calculation} - In regression tasks the output of each weak model is a numeric value, and the different outputs are combined using weighted median.
In the label ranking case, the output of each weak model is a ranking.  
Moreover, in the boosting case, each weak model has a different importance as measured by its weight.
Therefore we employed a weighted Borda method that is able to aggregate a set of rankings into a single ranking, and take into account their corresponding weights.
This aggregation step is formally described in equation \ref{eq:weighted_borda}.

\end{itemize}

%\subsection{LRBoost algorithm}\label{ss:lrboost_algorithm}
% Our algorithm is structured as follows: $LRBoost$ consists of iteratively learning base $LRT$ algorithms with respect to a distribution and adding them to a final strong model. 
% When they are added, they are weighted using the base learners' accuracy. In our case, accuracy calculation is done using Kendall-Tau coefficient function (details in equation \ref{eq:loss_by_kt} and in algorithm \ref{alg: LRBoost}). After a basic LRT learner is added, the data weights are re-weighting. By using KT similarity, input data whose predicted ranking is significantly different from the real ranking gain a higher weight than examples whose predicted ranking similar to their real ranking. Thus, future basic LRT learners focus more on the examples that previous basic learners predicted poorly. Finally, LRBoost algorithm combines the rankings by a Weighted Borda method that takes into account the accuracy weights of each base model. 

\section{Experimental Setting}
\label{sec:experimental_setting}

In this section, we describe the compared methods, the datasets used and the experimental flow.

\subsection{Compared Methods}

We compared the prediction performance of a simple \textbf{single model} for label ranking based on $LRT$, and the following four label ranking ensembles:

\begin{itemize}

\item \textbf{Boosting} - the $AdaBoost.LR$ algorithm for label ranking tasks that we developed.

\item \textbf{Bagging - Modal Ranking} - an existing label ranking ensemble which was described in \cite{aledo2017tackling}. 
This algorithm uses Bagging as the data sampling technique and Modal Ranking as the aggregation technique.

\item \textbf{Bagging - VRS} - an existing label ranking ensemble which was described in \cite{werbin2019beyond}.
This algorithm uses Bagging as the data sampling technique and a voting rule selector, which automatically selects the best voting rule to be used as the aggregation technique.

\item \textbf{Random Forest} - an existing label ranking ensemble which was described in \cite{sa2017label,ZHOU2018}.
This algorithm uses Random Forest as the data sampling technique and Borda as the aggregation technique.

\end{itemize}

To ensure a fair comparison, we re-implemented the compared ensembles, and all ensemble implementations used $LRT$ as the base label ranking algorithm.
$LRT$ was chosen as the base label ranking algorithm, since it was shown to work well in previous studies \cite{aledo2017tackling,hullermeier2008label}.

\subsection{Datasets}

As a benchmark, we used a total of 24 datasets.
The first 16 were proposed in \cite{cheng2009decision} and used since then as a standard benchmark for the label ranking problem.
They can be considered semi-synthetic (SS), as they were obtained by transforming a multi-class or regression problem to a label ranking one (see \cite{cheng2009decision} for the transformation details).
The last eight are real-world (RW) datasets.
Five of them correspond to real-world biological data problems and were used by \cite{hullermeier2008label}.
The labels in these datasets represent the expression level of genes of the yeast genome, where each gene has a phylogenetic profile of length 24 (i.e., 24 attributes).
In this case the expression profile of the output gene was directly converted into a rank (e.g. $(1.7, 2.9, 0.3, -2.4)$ was converted into $(2, 1, 3, 4)$).
The other three datasets were used by \cite{de2018discovering,werbin2019beyond}.
One dataset contains demographics and preferences over sushi types \cite{kamishima2003nantonac} and the other two contain social-economic and electoral results from Germany \cite{boley2013one}.

\begin{table}[t]
\caption{Datasets Characteristics}
\label{Table:datasets}
\begin{tabular}{@{}lllll@{}}
\toprule
\begin{tabular}[c]{@{}l@{}}
Dataset\end{tabular} & Type & Instances & Features & Labels \\ \midrule
Authorship  & SS          & 841     & 70 & 4 \\
Bodyfat     & SS          & 252     & 7  & 7 \\
Calhousing  & SS         & 20640   & 4  & 4 \\
Cpu-small   & SS          & 8192    & 6  & 5 \\
Elevators   & SS          & 16599   & 9  & 9 \\
Fried       & SS         & 40760   & 9  & 5 \\
Glass       & SS          & 214     & 9  & 6 \\
Housing     & SS          & 506     & 6  & 6 \\
Iris        & SS          & 150     & 4  & 3 \\
Pendigits   & SS         & 10992   & 16 & 10\\
Segment     & SS         & 2310    & 18 & 7 \\
Stock       & SS         & 950     & 5  & 5 \\
Vehicle     & SS          & 846     & 18 & 4 \\
Vowel       & SS          & 528     & 10 & 11\\
Wine        & SS          & 178     & 13 & 3 \\
Winsconsin  & SS          & 194     & 16 & 16\\
Cold        & RW              & 2465    & 24 & 4 \\
Diau        & RW              & 2465    & 24 & 7 \\
Dtt         & RW              & 2465    & 24 & 4 \\
Heat        & RW              & 2465    & 24 & 6 \\
Spo         & RW              & 2465    & 24 & 11\\
German2005  & RW              & 402     & 31 & 5 \\
German2009  & RW              & 407     & 33 & 5 \\
Sushi       & RW             & 5000    & 10 & 10\\ 
\bottomrule
\end{tabular}
\end{table}

\subsection{Experimental Flow}
Our experimental flow follows a similar approach to that of Aledo et al. \cite{aledo2017tackling} and Werbin et al. \cite{werbin2019beyond} and is composed of four main steps:% which are illustrated in Figure \ref{figure:flow}: 

\begin{enumerate}

\item \textbf{Cross Validation.} Given a label ranking dataset, we employed a 10-fold cross-validation process, in which the dataset was partitioned into ten distinct folds, and then in ten iterations, one fold was used as a test set and the remaining were used for the training set.

\item \textbf{Data Sampling and Training.} Given a training set, multiple bags were sampled, and each such bag was used to train a single label ranking prediction model.
For our evaluation, we used 50 bags.
In the case of boosting, the sampling of bags was done sequentially, where the sampling of a bag in a given iteration depended on the weights of training instances, which in turn depended on the performance of the weak model trained in the previous iteration.
%This dependency is illustrated in Fig. \ref{figure:lrboost_flow}. 

\item \textbf{Prediction.} Each one of the test instances was given as input to each one of the 50 trained models, and the 50 outputted rankings were aggregated into a single aggregated output, using Modal Ranking in the case of Bagging, Borda in the case of Random Forest and weighted Borda in the case of Boosting.

\item \textbf{Evaluation.} The aggregated ranking was compared to the real ranking using the Kendal-tau-b coefficient.
The results were averaged over all test instances, and then over the ten test folds, to produce a single measure for a dataset.

\end{enumerate}

% \begin{figure}[H]
% \centering
% \includegraphics[width=1.0\textwidth]{figures/flow.pdf}
% \caption{The general experimental flow}
% \label{figure:flow}
% \end{figure}

% \begin{figure}[H]
% \centering
% \includegraphics[width=1.0\textwidth]{figures/lrboost_flow.pdf}
% \caption{Changes to the experimental flow in the case of boosting}
% \label{figure:lrboost_flow}
% \end{figure}

%Steps A and B of the evaluation were implemented in Java 9.0.1, whereas the rest of the steps were implemented in Python 2.7.

\section{Results}
\label{sec:results}

In the first experiment, we wanted to compare the prediction performance of the five compared methods: Boosting, Random Forest, Bagging (two variations) and Single Model.
To do so, we first calculated the average Kendall-Tau-b for each method (over all instances in a given dataset). 
Then, we calculated the improvement in percentages in comparison to the simple model approach, for each dataset.
Finally, we averaged these numbers over all 24 datasets.

\begin{figure}[t]
\centering
\includegraphics[width=0.45\textwidth]{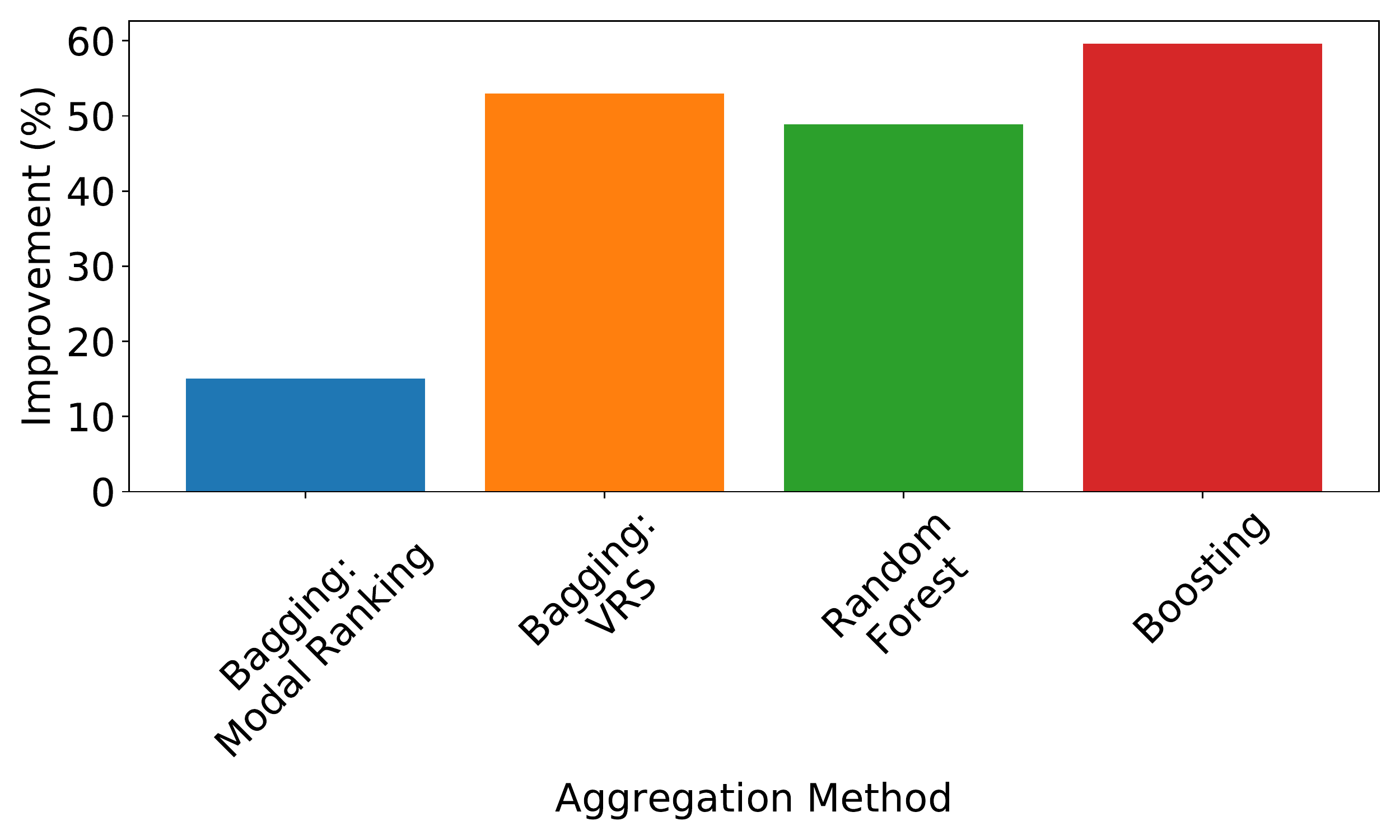}
\caption{Improvement of each ensemble method in comparison to the single model method}
\label{figure:fig1b}
\end{figure}

Fig. \ref{figure:fig1b} reports the average improvement (in percentages) of each of the ensemble methods in comparison to the single model approach.
As can be seen from the figure, all four ensemble methods performed better than the single model approach.
This result emphasizes the superiority of ensemble methods over simple models in general, and in the case of label ranking tasks in particular.
We further observe that Boosting outperforms the other ensemble methods with an average improvement of 60\% (compared to 49\% , 53\%, and 15\% obtained by Random Forest, Bagging: VRS and Bagging: Modal Ranking respectively).

\begin{figure}[t]
\centering
\includegraphics[width=0.45\textwidth]{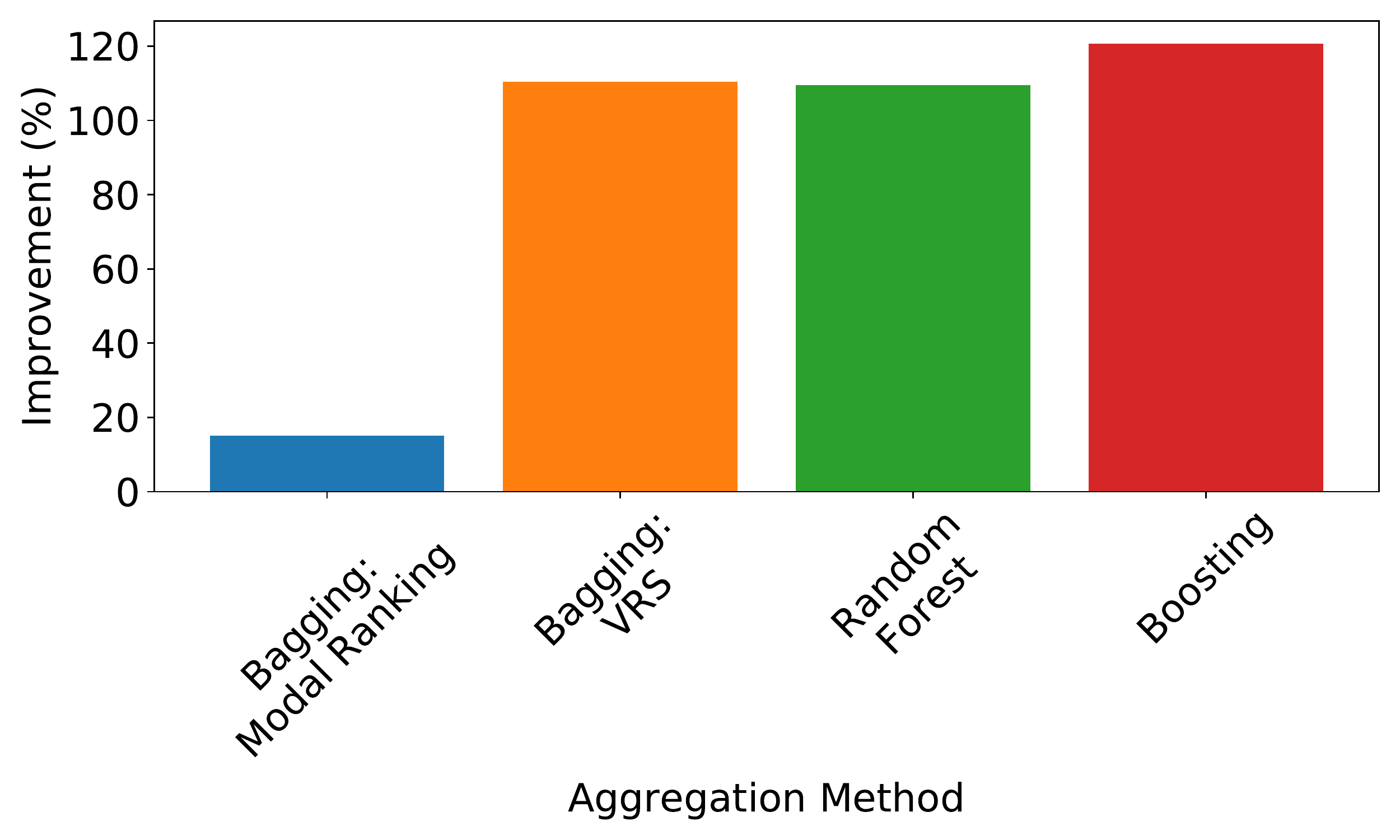}
\includegraphics[width=0.45\textwidth]{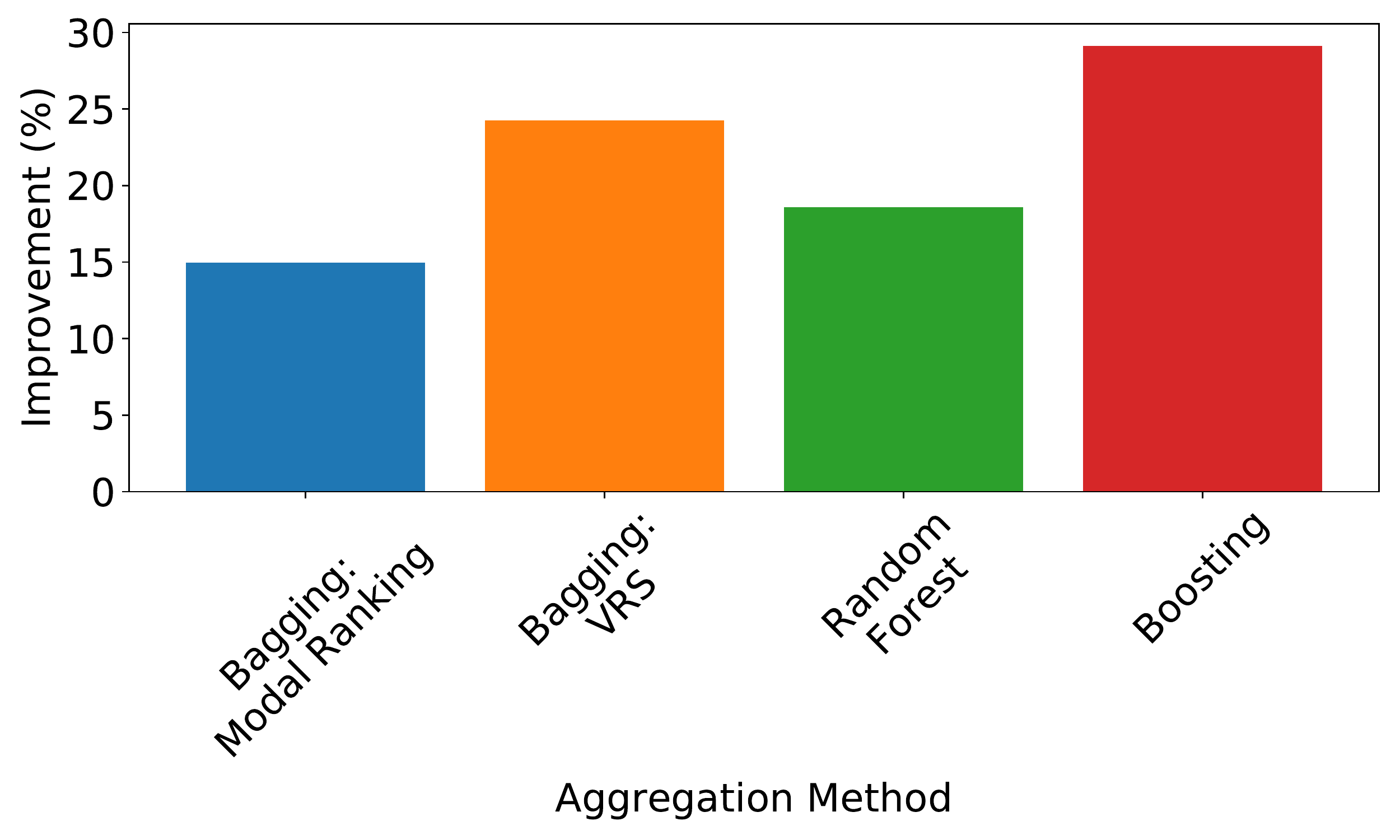}
\caption{Breakdown of improvement to real-world datasets (top) and semi-synthetic ones (bottom).}
\label{figure:fig1b_comp}
\end{figure}

Fig. \ref{figure:fig1b_comp} provides a breakdown of the results reported in Fig. \ref{figure:fig1b} to real-world datasets (left) and semi-synthetic ones (right).
As can be seen in the figure, the improvement obtained by all ensemble methods in the case of real-world datasets is substantially better.
For example, in the case of Boosting we observe an average improvement of 121\% in the case of real-world datasets, compared to an average improvement of 29\% in the case of semi-synthetic datasets.
%\todo{I think the following should be removed, since VRS is the 2nd best here}As a side note, we observe that while in the case of semi-synthetic datasets, Bagging with Modal Ranking and Random Forest perform roughly the same, in the case of real-world datasets, Random Forest manages to obtain a meaningful gap (110\% compared to 15\%).

\begin{figure*}[t]
\centering
\includegraphics[width=1.0\textwidth]{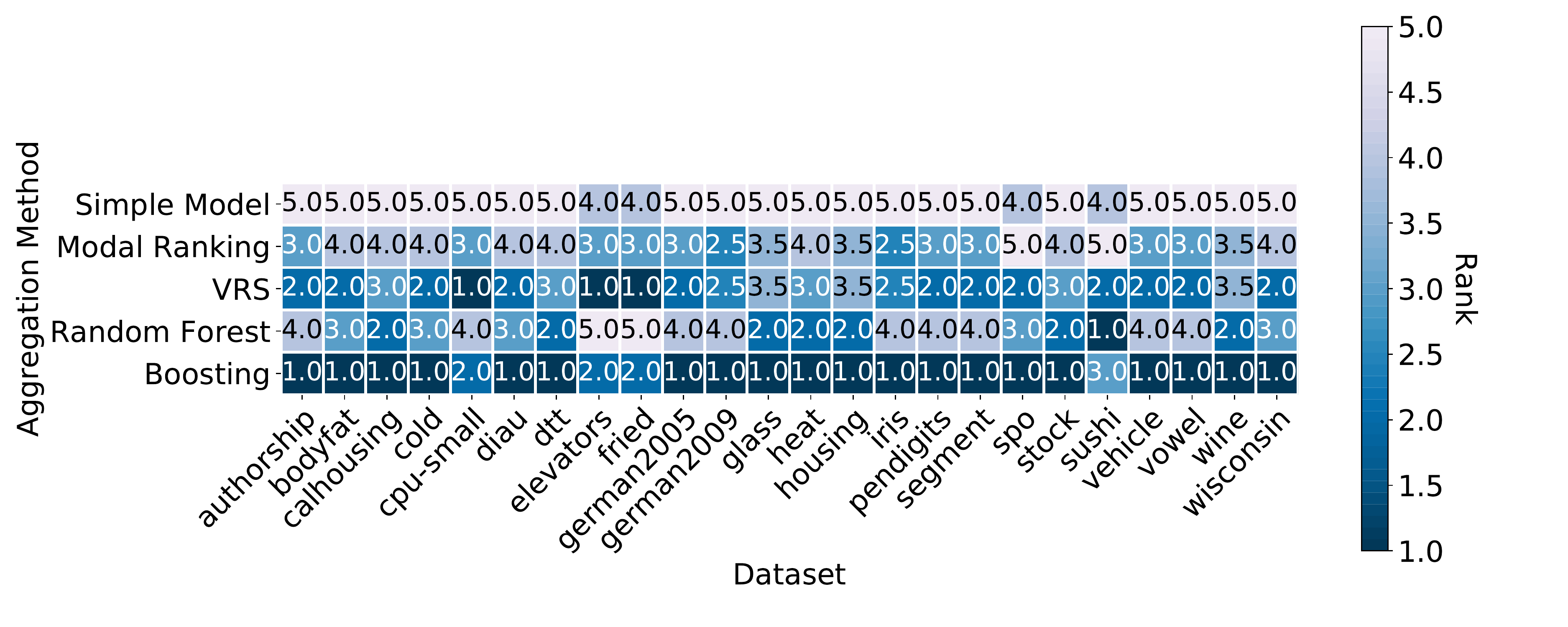}
\caption{Rank of each method for each of the 21 dataset}
\label{figure:fig6}
\end{figure*}

While the figures above reported the average results over multiple datasets, Fig \ref{figure:fig6} provides a breakdown to the 21 datasets, where each entry represents the rank of the given method for the given dataset (a rank of 1 means that the given method performed the best for the given dataset, and a rank of 4 means that it performed the worst).
As can be seen from Fig \ref{figure:fig6}, Boosting defeats the four other methods in 20 out of 24 of the datasets. 
In contrast, VRS was ranked first in only three of the 21 datasets, Random forest was ranked first once and for all 24 datasets - neither Bagging with Modal Ranking nor Simple Model obtained the first place.

% \begin{figure}[H]
% \centering
% \includegraphics[width=0.7\textwidth]{figures/fig3.pdf}
% \caption{XXX}
% \label{figure:fig3}
% \end{figure}

\begin{figure}[t]
\centering
\includegraphics[width=0.45\textwidth]{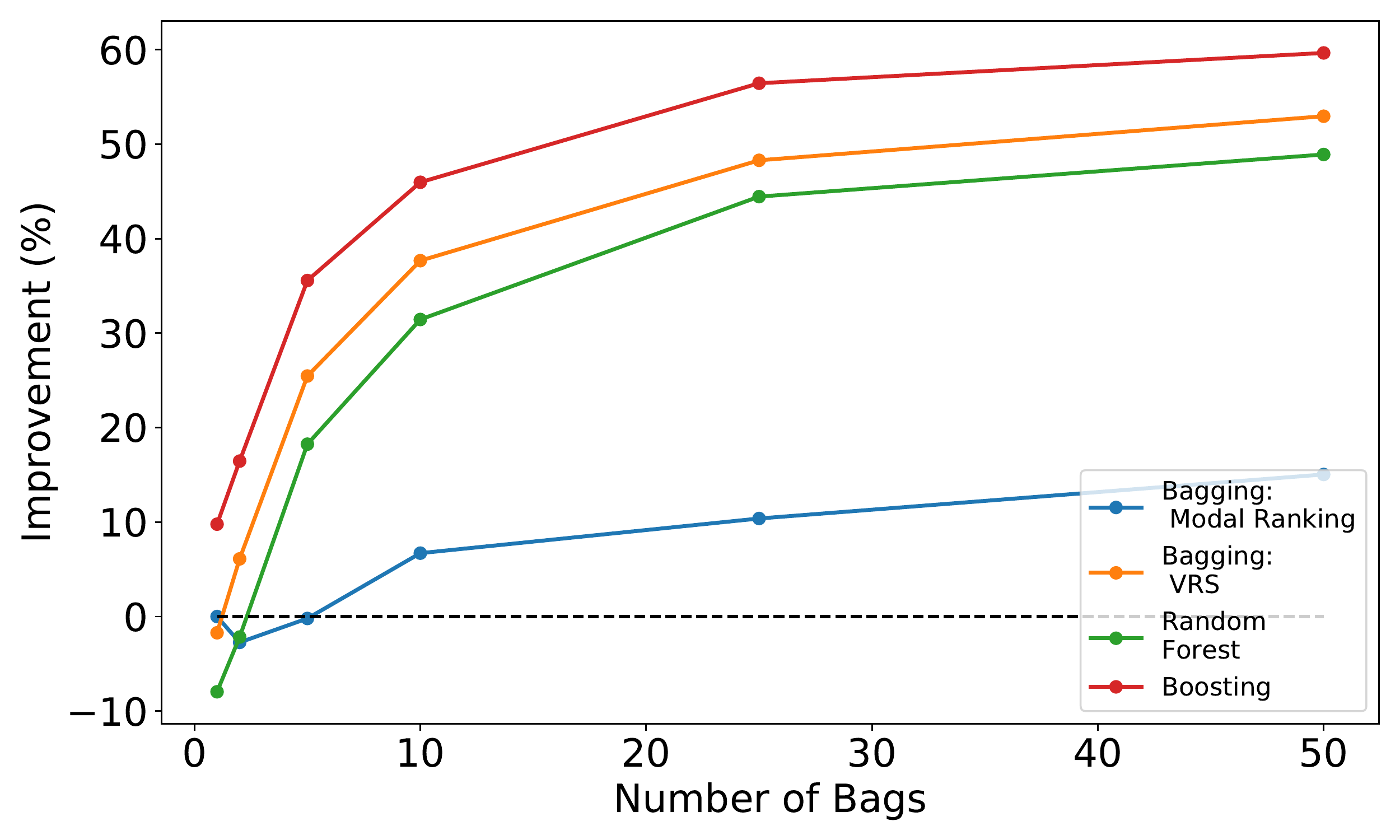}
\caption{Improvement as a function of the number of weak models used}
\label{figure:fig7}
\end{figure}

Finally, we were interested in testing how the number of weak models used in the ensemble affects prediction performance.
Fig \ref{figure:fig7} shows the improvement in percentages as a function of the number of weak models used for each of the aggregation methods.
As can be seen in Fig \ref{figure:fig7}, the prediction performance of all ensemble methods increases with the number of weak models used.
Moreover, we observe that the Boosting method outperforms the rest of the methods for all considered numbers of weak models.
Lastly, we notice a diminishing effect, where prediction performance seems to stabilize for all of the considered aggregation methods, when using 50 weak models.

In order to complete the picture, we performed statistical significance tests to compare between the five methods of interest.
We followed the robust non-parametric procedure described in \cite{garcia2010advanced}.
First, we used the Friedman Aligned Ranks test in order to reject the null hypothesis
that all methods perform the same. 
The Friedman Aligned Ranks test with a confidence level of 95\% rejected the null-hypothesis that all methods perform the same.
Then, we used the Conover post hoc test to perform pairwise comparison between the five methods.
We found that in accordance with our findings above,
Boosting performed significantly better than the other four methods.

\section{Conclusions}
\label{sec:conclusions}

In this paper, we proposed a novel boosting-based algorithm, AdaBoost.LR, for improving the prediction performance of label ranking ensembles.
An extensive evaluation that we performed showed that the proposed algorithm, performed significantly better than existing state-of-the-art algorithms for label ranking.
In particular, our algorithm defeated the benchmark methods in 20 out of the 24 publicly available label ranking datasets.

%We identified a few interesting future research directions.
%One direction concerns the choice of the base label ranking algorithm.
%In this paper we used LRT \cite{cheng2009decision}, but numerous label ranking algorithms were suggested in the literature.
%To the best of our knowledge, an extensive comparison of these algorithms in the context of ensembles is lacking.

%\todo{add one more direction - formal proof?}

%A closely related research direction would be to explore the case where incomplete rankings.
Our evaluation considered the case of complete rankings only.
However, given a base label ranking algorithm that accepts partial rankings as inputs (See for example \cite{aledo2017tackling}), AdaBoost.LR can be applied without any change.
Therefore, in future work, we plan to evaluate the performance of AdaBoost.LR in the case of partial rankings.

\flushcolsend

\newpage
It would be interesting to explore other types of label ranking ensembles in addition to the ones already studied (i.e., Bagging, Random Forest, and now Boosting).
For example, the Stacking approach \cite{wolpert1992stacked} can be extended to support label ranking tasks, by developing a meta label ranker that receives the output rankings of simple models as input attributes and returns a single aggregated ranking as output.

We believe that insights gained in the process of improving the competence of label ranking ensembles, may be valuable to a variety of fields that are concerned with rankings, including: combining voters' preferences in computational social choice \cite{chevaleyre2007short}, ensembles in machine learning, rank aggregation in information retrieval \cite{dwork2001rank}, group recommendations in recommender systems, phylogenetic profiling \cite{balasubramaniyan2004clustering}, and even error correction codes \cite{procaccia2016voting}.

\bibliographystyle{named}
\bibliography{references/refs}

\end{document}